\documentclass[letterpaper, 10 pt, conference]{ieeeconf}  

\IEEEoverridecommandlockouts                              

\usepackage{graphicx} 
\usepackage{blindtext}
\usepackage{todonotes}
\usepackage{bm}
\usepackage[nolist]{acronym}
\usepackage{amsmath} 
\usepackage{amssymb}  
\usepackage{capt-of} 
\usepackage{multirow}
\usepackage{mathabx}
\usepackage{tikz}

\acrodef{mpc}[MPC]{model-predictive control}
\acrodef{dnn}[DNN]{deep neural network}
\acrodef{cnns}[CNNs]{convolutional neural networks}
\acrodef{gb}[GB]{gallbladder}
\acrodef{iou}[IoU]{Intersection over Union}
\acrodef{ras}[RAS]{robot assisted surgery}
\acrodef{relu}[ReLU]{Rectified Linear Unit}
\acrodef{bce}[BCE]{binary crossentropy}

\newcommand\copyrighttext{%
  \footnotesize \textcopyright 2022 IEEE. Personal use of this material is permitted.
  Permission from IEEE must be obtained for all other uses, in any current or future 
  media, including reprinting/republishing this material for advertising or promotional 
  purposes, creating new collective works, for resale or redistribution to servers or 
  lists, or reuse of any copyrighted component of this work in other works.}
\newcommand\copyrightnotice{%
\begin{tikzpicture}[remember picture,overlay]
\node[anchor=south,yshift=10pt] at (current page.south) {\fbox{\parbox{\dimexpr\textwidth-\fboxsep-\fboxrule\relax}{\copyrighttext}}};
\end{tikzpicture}%
}

\title{\LARGE \bf
LapSeg3D: Weakly Supervised Semantic Segmentation of Point Clouds Representing Laparoscopic Scenes
}

\author{Benjamin Alt$^{1}$, Christian Kunz$^{2}$, Darko Katic$^{1}$, Rayan Younis$^{3}$, Rainer Jäkel$^{1}$,\\Beat Peter Müller-Stich$^{3}$, Martin Wagner$^{3}$, and Franziska Mathis-Ullrich$^{2}$
\thanks{*This  work  was  supported  by  the  German  Federal  Ministry  of  Education and Research under the grant 13GW0471C.}
\thanks{$^{1}$B. Alt, D. Katic and R. Jäkel are with Artiminds Robotics GmbH, 76131 Karlsruhe, Germany}%
\thanks{$^{2}$ C. Kunz and F. Mathis-Ullrich are with the Institute for Anthropomatics and Robotics, Karlsruhe Institute of Technology, 76131 Karlsruhe, Germany, 
{\small corresponding author: \tt franziska.ullrich@kit.edu}}%
\thanks{$^{3}$ R. Younis, B.P. Müller-Stich and M. Wagner are with the Department for General, Visceral and Transplantation Surgery, Heidelberg University Hospital, 69120 Heidelberg, Germany.}%
}

\begin{document}

\maketitle
\copyrightnotice
\thispagestyle{empty}
\pagestyle{empty}

\begin{abstract}
The semantic segmentation of surgical scenes is a prerequisite for task automation in robot assisted interventions. We propose LapSeg3D, a novel DNN-based approach for the voxel-wise annotation of point clouds representing surgical scenes. As the manual annotation of training data is highly time consuming, we introduce a semi-autonomous clustering-based pipeline for the annotation of the gallbladder, which is used to generate segmented labels for the DNN.  When evaluated against manually annotated data, LapSeg3D achieves an F1 score of 0.94 for gallbladder segmentation on various datasets of ex-vivo porcine livers. We show LapSeg3D to generalize accurately across different gallbladders and datasets recorded with different RGB-D camera systems.
\end{abstract}

\section{INTRODUCTION}
The understanding of the surgical scene is a crucial requirement of active robotic assistance and task automation in \ac{ras}. An important step in this pipeline is the semantic segmentation of the laparoscopic image frame, where every pixel or voxel is assigned a class label of the structure it belongs to, e.g. the gallbladder. 
In the standard of care, mono laparoscopes are used and the surgical scene is visualized to the surgeon on a monitor. In recent years, the use of stereo laparoscopes has increased, as they have become the state of the art in robotic telemanipulators, such as the da Vinci (Intuitive Surgical Inc., USA), improving the surgeon's spatial understanding of the surgical scene. However, their potential is not fully utilized.
In particular and towards autonomous robotic assistance in surgery, stereo laparoscopes offer several advantages to automated systems and navigation systems (with augmented reality or otherwise), as it is possible to reconstruct the three-dimensional surgical scene from the left and right image streams. This provides a better basis for surgical planning and the automated robotic execution of a task, e.g. grasping and tissue manipulation. To make specific use of the 3D-reconstructed surgical scene for surgical navigation, the individual points need to be semantically segmented, i.e. assigned to their corresponding tissue type (e.g. ``liver'' or ``gallbladder''). This is especially challenging due to the circumstances in \ac{ras}, such as the necessity for small sensors (which fit in the laparoscope), wet surfaces, blood and smoke in the scene, and oftentimes similar textures and colors of organs.

In current research, semantic segmentation of the surgical scene has primarily been focused on the recognition and pose estimation of surgical instruments in two-dimensional laparoscopic image frames \cite{bodenstedt2018comparative, bouget2017vision}, which is an essential step for skill automation \cite{chen2016virtual}. Presented approaches mainly utilize random forests \cite{bodenstedt2016image, allan2012toward} or \ac{cnns} \cite{pakhomov2019deep, garcia2016real, agustinos20152d, hasan2019u}.

\begin{figure}
    \centering
    \includegraphics[width=\linewidth]{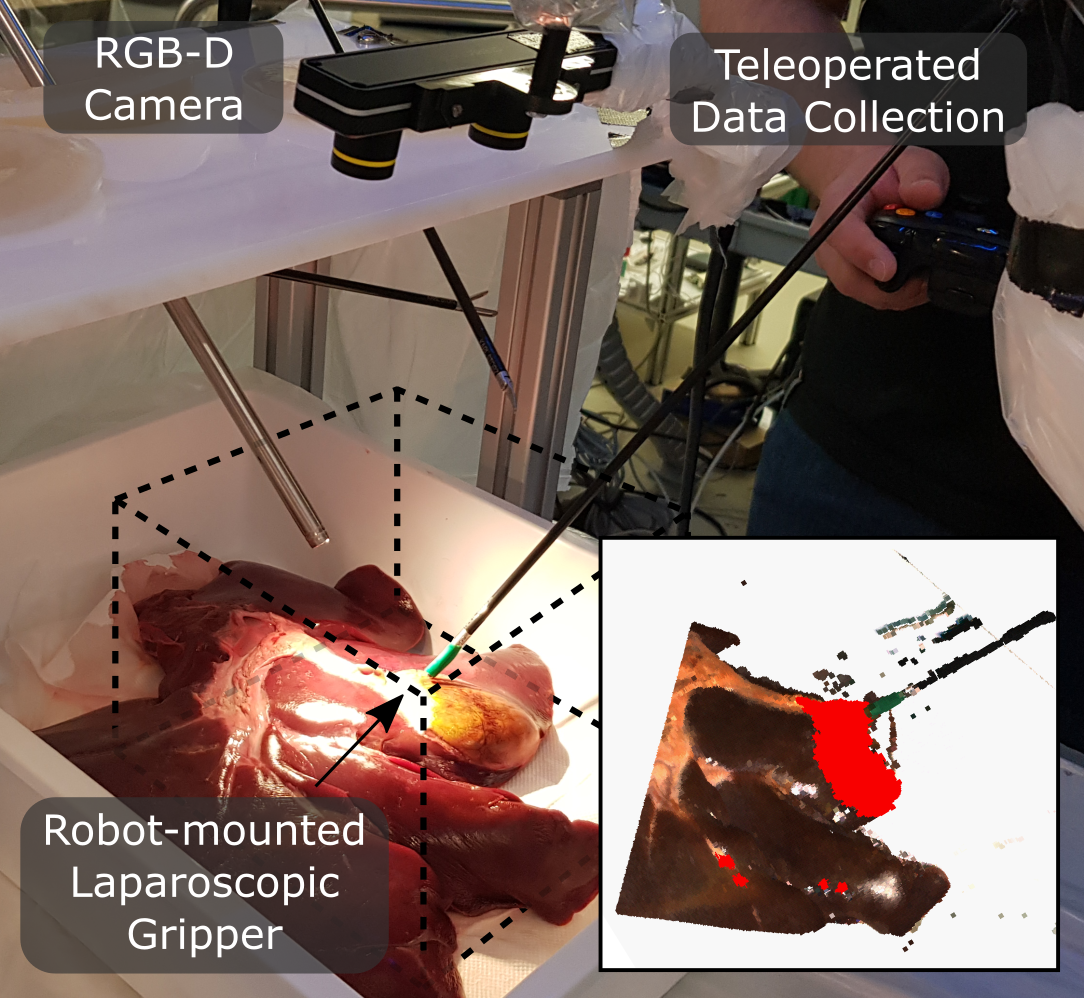}  
    \caption{Setup of the ex-vivo experiments and the resulting segmented point cloud of the surgical scene.}
    \label{fig:first_page_illustration}
\end{figure}




Few presented methods address the semantic segmentation of the full surgical scene.
Scheikl et al. \cite{scheikl2020deep} investigated the performance of different deep learning architectures (e.g. U-Net, TernausNet, FCN, LinkNet, SegNet) for the pixel-wise semantic segmentation of the complete surgical scene. Five classes were segmented (instruments, liver, gallbladder, fat and other). The structures were segmented with an \ac{iou} of 0.79.
Maqbool et al. \cite{maqbool2020m2caiseg} propose a method for pixel-wise semantic segmentation of the surgical scene by using a CNN Encoder-Decoder architecture. Additional works have been introduced as results of the MICCAI 2021 Endoscopic Vision `HeiSurf-Subchallenge' for surgical workflow analysis and full scene segmentation  \cite{bodenstedt_endoscopic_2021}.

\begin{figure*}
    \centering
    \includegraphics[width=\textwidth]{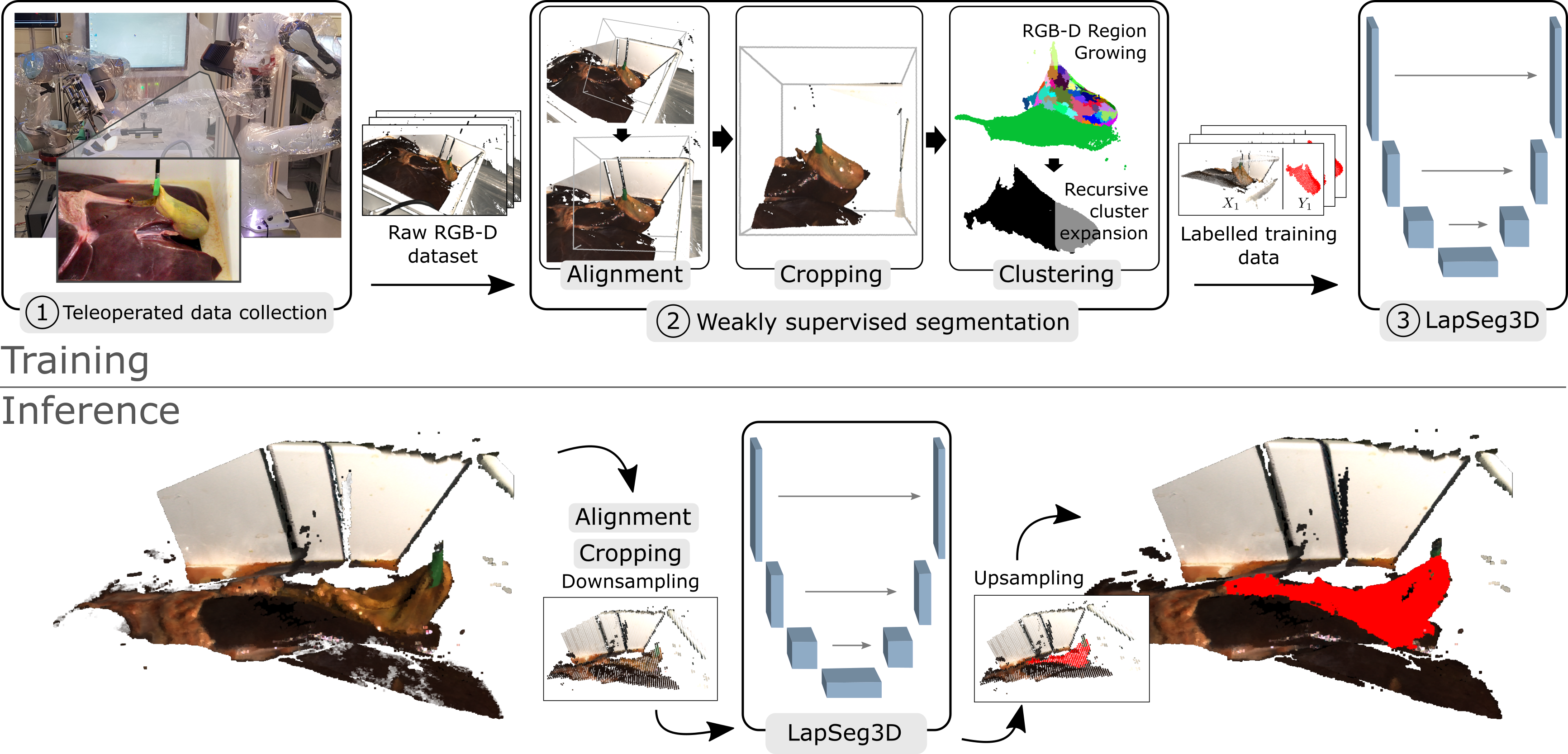}
    \caption{Overview of the proposed pipeline for the segmentation of point clouds representing laparoscopic scenes.}
    \label{fig:overview}
\end{figure*}

While these methods address the semantic segmentation of two-dimensional image frames of mono laparoscopes, few works exist that present approaches to perform a voxel-wise classification on three-dimensional point clouds.
Haouchine and Cotin \cite{haouchine2016segmentation} segment point clouds of laparoscopic scenes based on curvatures and normals to approximate connected surfaces of organs, e.g. the liver. However, the organs' textures are not taken into account. 

In other non-surgical domains several works for 3D image segmentation have been proposed. 
Qi et al. introduce a \ac{dnn} to process point clouds for various 3D recognition tasks, such as image segmentation \cite{qi2017pointnet, qi2017pointnet++}. Other works deal with object recognition and shape completion in point clouds, e.g. 3D Shape Nets \cite{wu20153d} and voxel-based methods \cite{wu20153d, brock2016generative}. Tchapmi et al. propose SegCloud, an end-to-end 3D segmentation framework for point clouds \cite{tchapmi2017segcloud}.
The LapSeg3D architecture proposed in this work is comparable to 3D-UNet proposed by Cicek et al. \cite{cciccek20163d}, where it is utilized for kidney segmentation in volumetric confocal microscopic data.

In this work, we focus on cholecystectomy, i.e. the removal of the gallbladder from the liver. We introduce LapSeg3D, a novel approach for semantic segmentation of point clouds to identify the 3D-surface of a gallbladder. Our core contribution is a weakly supervised clustering-based pipeline capable of bootstrapping a large, diverse dataset of labelled training examples (cf. Fig. \ref{fig:overview}). This dataset is then used to train a \ac{dnn} to perform the segmentation task. By generating its training data using weak supervision, LapSeg3D benefits from the fast inference and good generalization capabilities of \acp{dnn} while avoiding the need for manually labelled training data.
We evaluate our approach on a diverse set of ex-vivo gallbladders. LapSeg3D is highly accurate, achieving F1 scores of 0.94, and generalizes well across different gallbladders. During run-time, the complete segmentation from raw point cloud to upsampled segmentation results is performed in 162 ms ($\sigma$=3 ms) on a NVIDIA GeForce 3080 Ti GPU, with the \ac{dnn} itself performing inference in 17 ms ($\sigma$=1 ms). This enables intraoperative online usage, such as for navigation or visual servoing of robotic manipulators.

\section{METHODS}

\subsection{\ac{dnn} for RGB-D gallbladder segmentation}
\label{sec:segmentation_net}
\begin{figure}
    \centering
    \includegraphics[width=\linewidth]{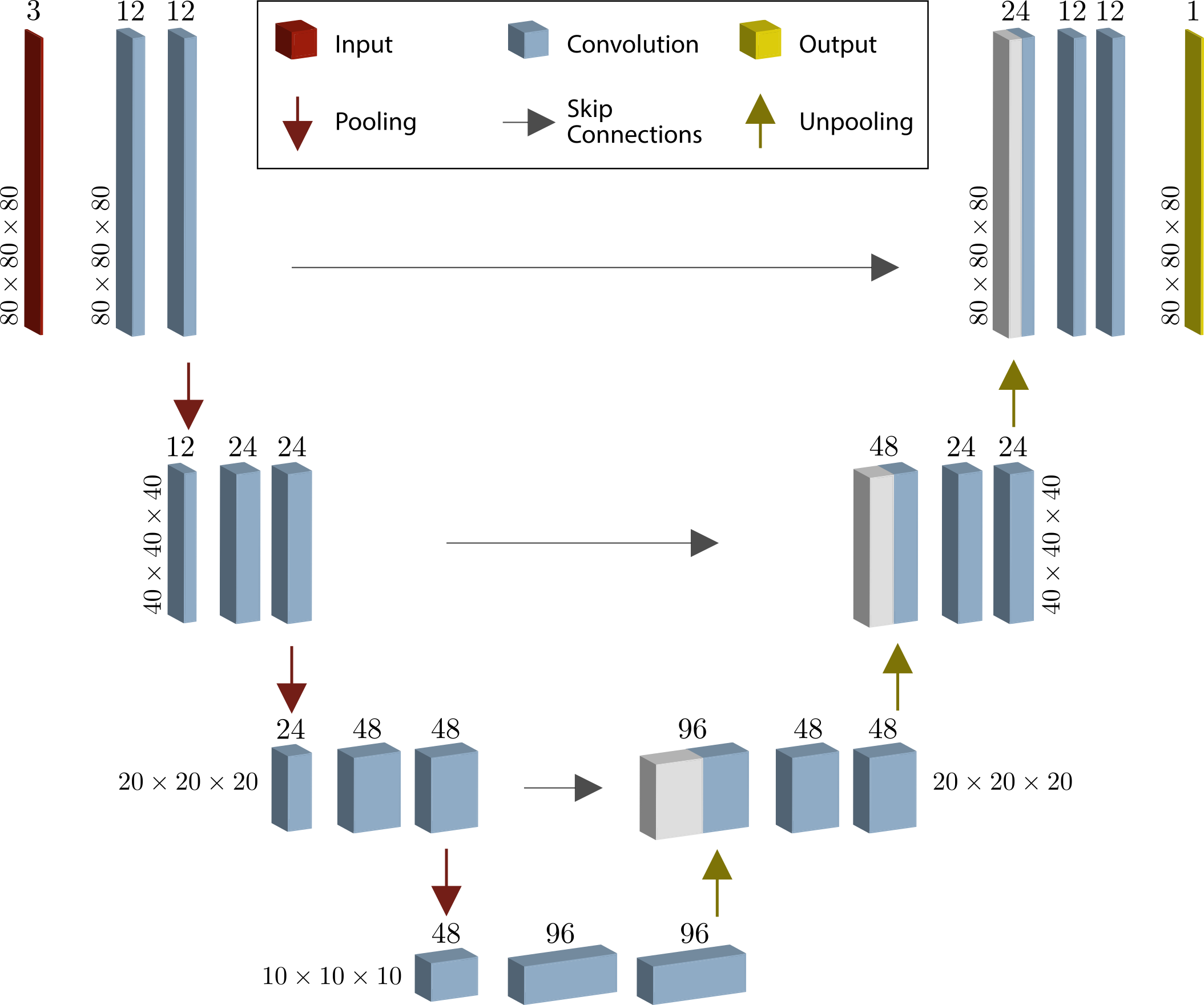}
    \caption{Neural network architecture of LapSeg3D.}
    \label{fig:unet_architecture}
\end{figure}

To segment the gallbladder in RGB-D images of surgical scenes, we propose a \ac{dnn} based on 3D-UNet \cite{cciccek20163d}. 3D-UNet is a 3D extension of UNet, a state-of-the-art architecture for 2D biomedical image segmentation \cite{alzahrani_biomedical_2021}. Our 3D-UNet implementation takes a 80x80x80 voxel grid $X$ of the surgical scene as input, where each voxel has three channels corresponding to the RGB color of the scene at this voxel. The network outputs a 80x80x80 voxel grid $\hat{Y}$ of the surgical scene, where each voxel has one binary channel indicating whether it is part of the gallbladder. The network architecture is shown in Fig. \ref{fig:unet_architecture}. There are several deviations from the original 3D-UNet architecture. Most notably, our network has much fewer parameters (746,365 compared to 19,069,955), due to the smaller dimensions of the input voxel grid as well as smaller feature maps at each layer. The smaller number of parameters reduces the hardware requirements at runtime: LapSeg3D has a memory footprint of 2700 MB, enabling our network to perform inference on CPUs and consumer-grade GPUs. We use instance normalization before each \ac{relu} activation, as we empirically found it to outperform batch normalization for our application. The weighted Softmax output activation function in \cite{cciccek20163d} was replaced by a Sigmoid activation, as the \ac{dnn} learns a voxel-wise binary classification task and labels are available for every voxel.

Given a training dataset $\mathcal{D}_{train} = \{(X_0, Y_0), ..., (X_N, Y_N)\}$ containing $N$ input-label pairs, we use the Adam optimizer \cite{kingma_adam_2017} to minimize the \ac{bce} between the output voxel grids $\hat{Y}$ and their corresponding ground-truth voxel grids $Y$:
\begin{equation}
    BCE(Y, \hat{Y}) = -\frac{1}{|Y|}\sum^{|Y|}_{i=1}\Big[Y^i\log(\hat{Y}^i)+(1-Y^i)\log(1-\hat{Y}^i)\Big],
\end{equation}
where $Y^i$ and $\hat{Y}^i$ denote the i\textsuperscript{th} entries in voxel grids $Y$ and $\hat{Y}$.
This corresponds to a supervised training regime, as the labelled voxel grids $Y$ are used during training.

In real-world surgical applications, any solution for segmenting surgical scenes must generalize to new gallbladders not present in the training dataset, as it is difficult to collect new patient-specific training data and re-train the network before surgery. We leverage data augmentation to improve the generalization capacity of the \ac{dnn}. Random rotation around the coordinate axes ($rx,ry,rz \in [-30^{\circ}, 30^{\circ}]$), uniform scaling (factor $s \in [0.8, 1.2]$) and elastic transformation (factor $f \in [0, 0.3]$) are applied with the probability $p=0.5$ to each input-label tuple per epoch. Random gamma correction ($\gamma \in [0.7, 1.5]$), contrast ($\alpha \in [0.7, 1.3]$), and brightness ($\beta \in [-0.3, 0.3]$) are always applied.

\subsection{Weakly supervised gallbladder segmentation}
\label{sec:weakly_supervised_segmentation}
The proposed segmentation \ac{dnn} is trained on labelled input-label pairs via a supervised learning regime. To ensure the applicability of the approach in real-world surgical settings and minimize human involvement during training, we propose to convert it into a weakly-supervised regime by generating high-quality labelled training data using a clustering-based pipeline.
To bootstrap a large training dataset for the segmentation \ac{dnn}, we contribute the multi-stage weakly supervised data processing pipeline outlined in Fig. \ref{fig:overview} (2). Its objective is to generate high-quality labelled training examples for the \ac{dnn} by segmenting the gallbladder from a large number of raw RGB-D images of surgical scenes, while requiring a minimal amount of human intervention.

The proposed pipeline is composed of three distinct stages (cf. Fig. \ref{fig:overview} (2)), which are described below. Human involvement is only required for the very first training example - all other RGB-D images in the dataset are processed autonomously.

\subsubsection{Alignment}
To ensure robustness against changes in the relative position of the camera with respect to the scene, the raw RGB-D point cloud is first \textit{aligned} to a reference scene $P_{ref}$ via a homogeneous transformation $T_{align}$. Our approach assumes a static camera pose per dataset, but allows for changing camera poses between datasets. Aligning all datasets to the same reference scene permits the network to be trained across many different surgical scenes covering multiple patients and camera perspectives. For surgical practice, this implies that the pose of the stereo laparoscope must not change during surgery, or that the current laparoscope pose must be known (i.e. tracked) at all times.
For our experiments, $T_{align}$ is computed via least-squares regression over point correspondences between the current and reference point clouds. For the first point cloud $P_1$ of a dataset, a human is asked to identify at least four point correspondences between $P_1$ and $P_{ref}$. 
In practice, we found that identifying points at the tip (fundus) as well as at the neck (infundibulum, collum) of the gallbladder in both $P_1$ and $P_{ref}$ produced sufficient alignment. The computed $T_{align}$ is used for the remainder of the dataset.

\subsubsection{Cropping} The aligned point cloud is \textit{cropped} to contain only the surgical scene. This is particularly important for cameras or stereo laparoscopes with a field of view much larger than the scene. Again, a human is asked to select a bounding box for $P_1$, which is then applied to the remainder of the dataset.

\subsubsection{Clustering-based gallbladder segmentation}
\label{sec:segmentation_region_growing}
Gallbladder segmentation is performed by an algorithm, which combines clustering with heuristics specific to the domain of organ segmentation. In a first step, color-based 3D region growing \cite{zhan_color-based_2009} is leveraged to split the raw point clouds into clusters of similar color. Similar to the alignment and cropping stages, a human expert provides a set of gallbladder colors by picking points in $P_1$. Clusters with colors sufficiently different from any picked gallbladder colors are rejected outright. The remaining clusters are merged into the final gallbladder cluster $C_{GB}$ via the following algorithm:
\begin{enumerate}
    \item $C_{GB}$ is initialized with the largest cluster.
    \item All clusters adjacent to $C_{GB}$ are added to $C_{GB}$. Two clusters are adjacent if the smallest distance between any two points in the clusters is below a given threshold.
    \item The process is iterated until $C_{GB}$ ceases to grow.
\end{enumerate}

The generated training dataset then consists of the aligned and cropped RGB-D images of the surgical scene as well as the corresponding segmentation results. Both input and label images are downsampled to an 80x80x80 voxel grid.


\section{EXPERIMENTAL VALIDATION}
\label{sec:experiments}

We conduct a series of experiments on ex-vivo porcine models to assess the validity of our approach.

\subsection{Experimental setup}
The experimental setup is illustrated in Fig. \ref{fig:first_page_illustration}. An operating table is equipped with a UR5 industrial manipulator (Universal Robots A/S, Odense, Denmark). The robot's end effector is a standard laparoscopic gripper (KARL STORZ SE \& Co KG, Tuttlingen, Germany), which is connected to the robot via a customized mechatronic interface. 
The segmentation of the instrument is not within the scope of this work. State-of-the-art algorithms achieve high F1 scores (approx. 0.88) for the semantic segmentation of the complete instrument \cite{bodenstedt2018comparative}. To easily track the instrument, we apply green markers to the gripper fingers. The surgical procedures are performed on ex-vivo porcine livers by a human surgeon. Two RGB-D cameras acquire data for the experiments: A Zivid One industrial RGB-D camera (Zivid AS, Oslo, Norway), operating via structured light and providing point clouds with high dynamic range at a resolution of 1920 x 1200 and a frame rate of up to 10 Hz; as well as a ZED Mini RGB-D camera (Stereolabs Inc., San Francisco, U.S.A), with lower dynamic range, but higher frame rate of 15 Hz at a resolution of 4416x1242 (left and right image side by side).

\subsection{Data collection} We collected a total of 15 datasets, each containing between 10 and 483 raw point clouds of the surgical site, from a total of nine ex-vivo porcine models. The datasets cover different stages of the operation from three different camera angles. Six datasets were recorded using the Zivid camera, while the Stereolabs camera was used for the remaining nine. Each dataset was collected using the following protocol:
\begin{enumerate}
    \item The liver is positioned upside-down compared to the in-vivo anatomy, i.e. with the gallbladder and the hilum of the liver facing upwards. Thus, the gallbladder is always visible. This results in a view similar to the intraoperative view when gallbladder and liver are elevated by a grasper at the gallbladder fundus.
    \item Blunt or electrocautery dissection of Calot's triangle; clipping and cutting of cystic duct and cystic artery. No RGB-D data is collected during this step.
    \item Teleoperated grasping of the gallbladder by the robot. The grasp point is dependent on the state of the operation. Initially, the gallbladder is grasped at the infundibulum. At later stages, the grasp point advances along the body toward the fundus to remain close to the respective dissection plane between liver and gallbladder.
    \item Execution of random teleoperated end-effector motions for 2-3 minutes. The teleoperator mimics the range and type of gripper motions (i.e., lateral, upward, and backward) commonly performed by the surgeon. RGB-D point clouds of the surgical site are continuously recorded during teleoperation.
    \item Teleoperated release of the gallbladder and continuation of gallbladder removal by the surgeon.
    \item Steps 2-4 are repeated respectively after the removal of the first, second and third fifths of the gallbladder.
\end{enumerate}
In the first experiments (Stereloabs 1-5, Zivid) the liver and gallbladder were positioned in a white box, which caused reflections of the gallbladder visible in the RGB-D images. Thus, for the last four datasets (Stereolabs 6-9), a matte black box was used. 

\subsection{Experiments}
\subsubsection{Comparison with expert labels}
\label{sec:comparison_expert_labels}
In a first series of experiments, we compare the segmentation results of LapSeg3D with a total of 67 RGB-D scenes manually segmented by surgeons using CloudCompare \cite{noauthor_cloudcompare_nodate}. Three models were trained on three different training datasets: all\textsubscript{RG} containing a total of 2,311 RGB-D images of surgical scenes from both cameras; sl\textsubscript{RG}, containing 1,921 images from the Stereolabs camera; and ziv\textsubscript{RG}, containing 390 images from the Zivid camera. Training times for the networks varied with the size of the dataset. For the largest dataset (all\textsubscript{RG}), training took approximately 8 hours on a NVIDIA 3090 GPU. All training data was generated using the region growing-based weakly supervised processing pipeline (RG). To assess the data generation pipeline, its outputs are also included in the analysis.

\begin{figure}
    \centering
    \includegraphics[width=\linewidth]{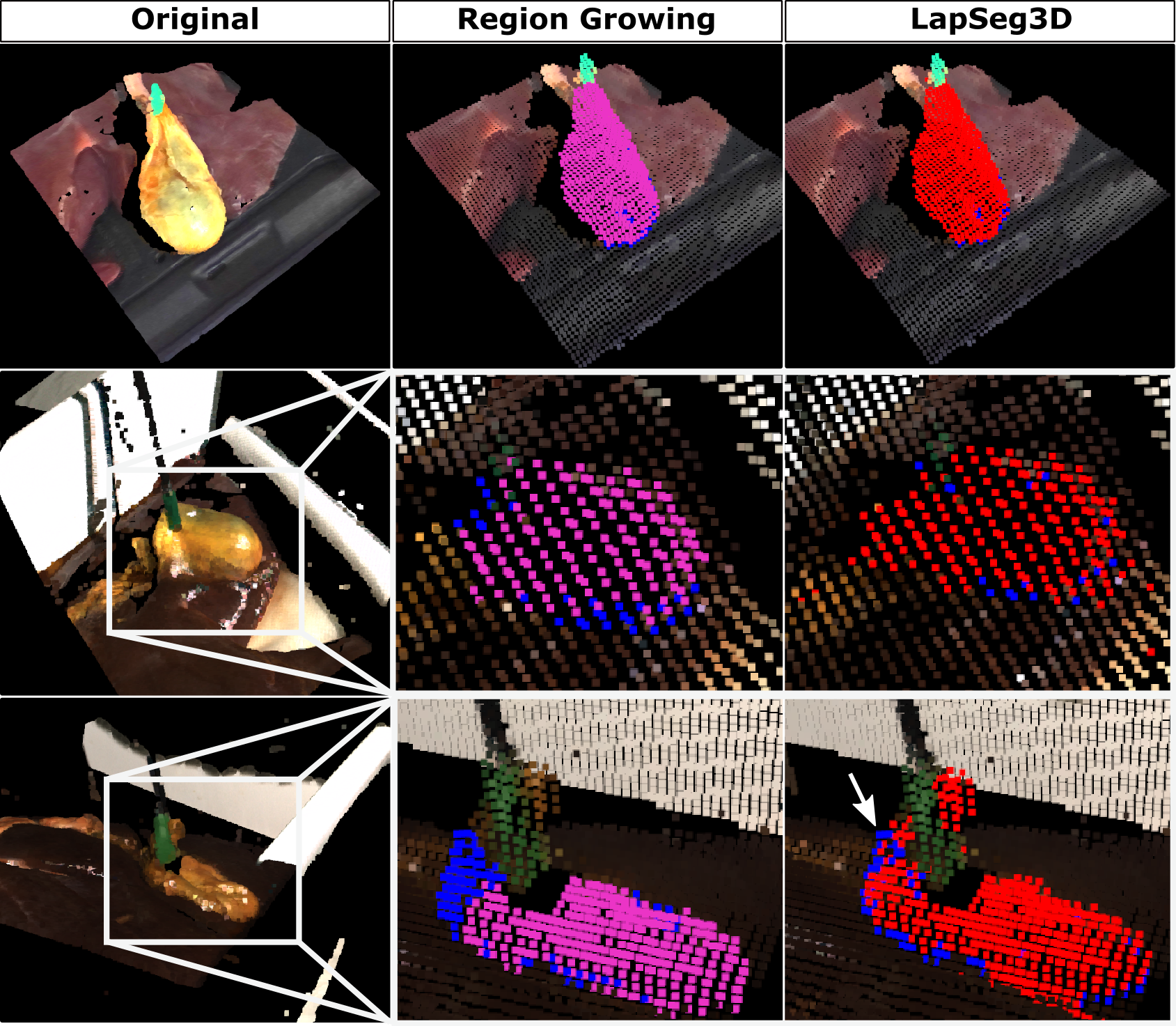}
    \caption{Segmentation results (from left to right): original point cloud recorded by sensor, annotation derived through region growing (magenta), annotation predicted through our proposed LapSeg3D (red), compared to manual annotations of the medical experts (blue).}
    \label{fig:detailed_example_pcds}
\end{figure}

\subsubsection{4-fold crossvalidation}
\label{sec:4fold_crossvalidation}
In a second series of experiments, we perform 4-fold crossvalidation to assess the capacity of LapSeg3D to generalize to scenes and gallbladders beyond the training dataset. Four networks were trained, each on a subset of the datasets collected with the Stereolabs camera: sl678\textsubscript{RG}, sl689\textsubscript{RG}, sl789\textsubscript{RG} and sl679\textsubscript{RG}, combinations of data from the 6\textsuperscript{th}, 7\textsuperscript{th}, 8\textsuperscript{th} and 9\textsuperscript{th} gallbladder. The networks are evaluated on the labels of the respective missing gallbladder provided by a human surgeon.

\section{RESULTS}

\begin{table}[]
 \renewcommand{\arraystretch}{1.3}
 \centering
     \caption{Performance of LapSeg3D compared to human expert labels.}
     \label{tab:snn_evaluation}
\begin{tabular}{lllrrrr}
\hline
 Model      & $\mathcal{D}_{train}$   &   $\mathcal{D}_{test}$     &   P &   R &   F1 &    IoU\\
\hline
 \textbf{LapSeg3D} & \textbf{all\textsubscript{RG}}                &            \textbf{all\textsubscript{H}}        &      \textbf{0.94} &   \textbf{0.95} &     \textbf{0.94} & \textbf{0.89} \\
  \textbf{LapSeg3D} &  \textbf{all\textsubscript{RG}}                &             \textbf{sl\textsubscript{H}} &       \textbf{0.95} &    \textbf{0.95} &      \textbf{0.95} &  \textbf{0.90} \\
  \textbf{LapSeg3D} &  \textbf{all\textsubscript{RG}}                &             \textbf{ziv\textsubscript{H}}      &       \textbf{0.91} &    \textbf{0.96} &      \textbf{0.93} &  \textbf{0.88} \\\hline
 LapSeg3D & sl\textsubscript{RG}                 &            all\textsubscript{H}        &      0.95 &   0.82 &     0.86 & 0.79 \\
 LapSeg3D & sl\textsubscript{RG}                 &            sl\textsubscript{H} &      0.97 &   0.91 &     0.94 & 0.89 \\
 LapSeg3D & sl\textsubscript{RG}                 &            ziv\textsubscript{H}      &      0.92 &   0.67 &     0.72 & 0.61 \\\hline
 LapSeg3D & ziv\textsubscript{RG}              &            all\textsubscript{H}        &      0.93 &   0.84 &     0.87 & 0.78 \\
 LapSeg3D & ziv\textsubscript{RG}              &            sl\textsubscript{H} &      0.91 &   0.80 &     0.84 & 0.73 \\
 LapSeg3D & ziv\textsubscript{RG}              &            ziv\textsubscript{H}      &      0.96 &   0.91 &     0.93 & 0.88 \\\hline
  Region Growing         & -                  &          all\textsubscript{H}        &      0.96 &   0.92 &     0.94 & 0.89 \\
 Region Growing          & -                  &       sl\textsubscript{H} &      0.96 &   0.92 &     0.94 & 0.88 \\
 Region Growing          & -                  &       ziv\textsubscript{H}      &      0.95 &   0.93 &     0.94 & 0.89 \\
\hline
\end{tabular}\vspace{4pt}
P: Precision, R: Recall, F1: F1 Score, IoU: Intersection over Union.\\
RG: Labels generated by weakly supervised region growing.\\
H: Labels annotated by a human surgeon.\\ 
sl: Stereolabs camera, ziv: Zivid One camera.
\end{table}

The results of experiment 1 (see Section \ref{sec:comparison_expert_labels}) are summarized in Table \ref{tab:snn_evaluation}. Trained on a dataset comprising all nine gallbladders, LapSeg3D achieves an F1 score of 0.94 and an \ac{iou} score of 0.89 when compared against labels provided by a medical expert. Echoing the results in \cite{bodenstedt_endoscopic_2021}, training on data from one camera (e.g. Zivid in ziv\textsubscript{RG}) and evaluating on data from another (e.g. Stereolabs in sl\textsubscript{H}) lowers F1 and \ac{iou} scores. Given the large differences in the dynamic range and resolution between the two cameras, however, the scores (F1 = 0.84, \ac{iou} = 0.73) are highly competitive, particularly given that the training set ziv\textsubscript{RG} contained only 390 images of three gallbladders. Segmentation of raw point clouds with LapSeg3D took a mean of 162 ms ($\sigma$=3 ms) per point cloud, with the \ac{dnn} itself requiring 17 ms ($\sigma$=1 ms) on an NVIDIA 3080 Ti GPU. The weakly supervised labelling pipeline also achieves F1 scores of 0.94 and \ac{iou} scores above 0.88 on all datasets, demonstrating its capacity to bootstrap high-quality labels with little human involvement. 

Qualitative analysis of the segmented point clouds confirms the results. Fig. \ref{fig:detailed_example_pcds} (top) shows a raw RGB-D point cloud from the Stereolabs camera (left), human-provided ground truth labels (blue) and the segmentation results of region growing (magenta) and LapSeg3D (red) in the downsampled voxel space, which both provide very good segmentation. The middle and bottom rows show segmentation results for Zivid data. In both cases, LapSeg3D provides accurate segmentation. The bottom row illustrates our observation that LapSeg3D segments even strongly deformed gallbladders well, including parts of the collum (marked in Fig. \ref{fig:detailed_example_pcds} with a white arrow), which region growing failed to segment.

\begin{figure}
    \centering
    \includegraphics[width=\linewidth]{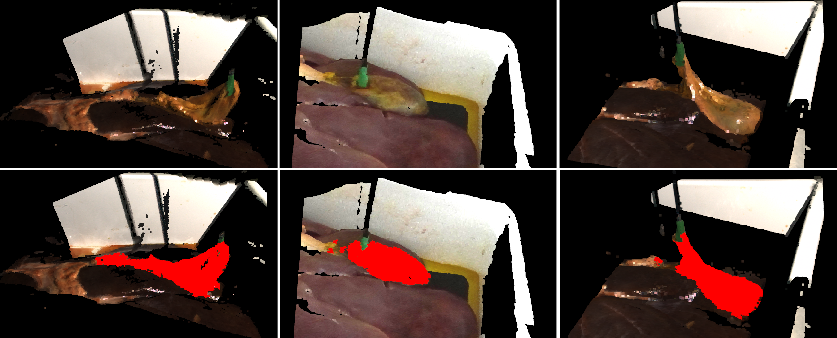}
    \caption{Results of different segmentations upsampled to their original size.}
    \label{fig:example_pcds}
\end{figure}

Fig. \ref{fig:example_pcds} shows examples from the test data in all\textsubscript{RG}, illustrating the capacity of LapSeg3D to segment gallbladders in different deformation states, during different phases of the operation, and grasped at different points. Unlike Fig. \ref{fig:detailed_example_pcds}, Fig. \ref{fig:example_pcds} shows the segmentation results upsampled to the original sensor resolution. 

Table \ref{tab:4fold_crossvalidation} summarizes results of 4-fold crossvalidation. LapSeg3D achieves an average F1 score of 0.84 and an average \ac{iou} of 0.74, indicating that it generalizes well to new scenes. The results are evidence that training data from three gallbladders suffice for LapSeg3D to learn sufficiently meaningful features to segment unseen gallbladders. Given that individual gallbladders vary greatly from patient to patient and gathering patient-specific data is costly, they indicate that LapSeg3D and its weakly supervised training scheme are well suited to real-world surgical applications.

\begin{table}[]
 \renewcommand{\arraystretch}{1.3}
 \centering
     \caption{Performance of LapSeg3D under 4-fold crossvalidation.}
     \label{tab:4fold_crossvalidation}
\begin{tabular}{lllrrrr}
\hline
 Model      & $\mathcal{D}_{train}$   &   $\mathcal{D}_{test}$   &   P &   R &   F1 &    IoU \\
\hline
LapSeg3D & sl678\textsubscript{RG}       &            sl9\textsubscript{H}          &      1.0 &   0.75 &     0.86 & 0.75 \\
LapSeg3D & sl679\textsubscript{RG}       &            sl8\textsubscript{H}           &      0.83 &   0.96 &     0.89 & 0.81 \\
LapSeg3D & sl689\textsubscript{RG}       &            sl7\textsubscript{H}           &      0.93 &   0.85 &     0.89 & 0.83 \\
LapSeg3D & sl789\textsubscript{RG}       &            sl6\textsubscript{H}           &      0.69 &   0.81 &     0.73 & 0.58 \\
\hline
\end{tabular}
\end{table}

\section{DISCUSSION}

\subsection{Weakly-supervised training pipeline}
We propose a pipeline for weakly-supervised generation of training data. While not fully unsupervised, human involvement during training is reduced to selecting points and bounding boxes on one image per dataset. Quality of the input depends on the user's experience and can lead to sub-optimal results, when non-representative points of the gallbladder are chosen. The weakly-supervised training pipeline achieves an F1 score of 0.94. Methods to sort out remaining bad training data are the subject of future work.


\subsection{LapSeg3D}
Other works have shown that it is challenging for a trained neural network to be used on image data from unknown sensor systems. When LapSeg3D is trained on the Stereolabs dataset, it still achieves a good F1 score of 0.72 when evaluated on Zivid point clouds. This rate can be further increased when the training dataset is enhanced with Zivid data. Future work will investigate how results on unknown camera systems can be further improved, e.g. through additional data augmentation.

An additional challenge in biomedical image segmentation is the great variance in patients' anatomies and pathologies, which can lead to different colouring, texture, and dynamic behaviour of the gallbladder. LapSeg3D copes well with these variations, achieving a highly competitive average F1 score of 0.84 on a 4-fold crossvalidation task.

Our approach is most challenged by different lighting conditions and noise in the point clouds. This can lead to errors segmenting the borders of the gallbladder and the transition to the liver, but only small areas around the gallbladder are affected.

We have shown that LapSeg3D is able to learn to segment structures that are not present in the training data, such as the infundibulum, which is grasped by a laparoscopic gripper, as depicted in Fig. \ref{fig:detailed_example_pcds} (white arrow). In future work, we will consider the detailed semantic segmentation of the gallbladder into its anatomical components, with dedicated labels for e.g. cystic duct, infundibulum and fundus.

The definition of whether voxels are part of the gallbladder may not be always clear and depend on the annotator. This mainly concerns structures such as the infundibulum, the collum and Calot’s triangle. Here, we define the gallbladder starting at the collum, but even for a medical expert this can vary in the manually annotated datasets.

With an overall F1 score of 0.94, our approach outperforms the state of the art for segmenting 2D images. This could be an indication that the 3D structure of an organ is also taken into account during segmentation.

\subsection{Limitations}
The acquired point clouds were recorded using two external camera systems. It can be assumed that the ZED mini camera is comparable to state-of-the-art stereo laparoscopes, as they provide a similar resolution. 3D reconstruction was not considered in this work. Both camera systems provide methods to automatically reconstruct the 3D scene from image data. Therefore, our approach needs to be transferred to laparoscopic sensors, such as stereo laparoscopes.
The ex-vivo livers were recorded at a distance of approximately 35 cm and were always visible at the center of the point clouds. When recording data inside the human body, the field of view may be limited. We have shown that our approach was able to learn the segmentation of difficult deformation states of the gallbladder. Future work will transfer LapSeg3D to laparoscopic image data with a limited field of view.

\section{CONCLUSION}
This work presents a method for fast and reliable gallbladder segmentation in point clouds of surgical scenes as a basis for automated robotic gallbladder removal. The presented neural network LapSeg3D was trained using a weakly supervised method for automated generation of training data utilizing a region growing approach, and was shown to be able to perform voxel-wise segmentation of laparoscopic scenes with an F1 score of 0.94 in 17 ms.

\addtolength{\textheight}{-12cm}   





\bibliographystyle{IEEEtran}
\bibliography{Literature}

\end{document}